\documentclass{ifacconf}

\usepackage{booktabs} 
\usepackage{bm} 
\usepackage{algorithm}
\usepackage{algpseudocode}
\usepackage{algorithmicx}
\usepackage{amsmath}
\usepackage{amssymb} 
\usepackage{subcaption}
\usepackage{graphicx}      
\usepackage{natbib}        
\begin{document}
\begin{frontmatter}

\title{Safe Heterogeneous Multi-Agent RL with Communication Regularization for Coordinated Target Acquisition\thanksref{footnoteinfo}} 

\thanks[footnoteinfo]{This work was partially supported by the Wallenberg AI, Autonomous Systems and Software Program (WASP) funded by the Knut and Alice Wallenberg Foundation. The authors are within the Robotics and AI Group, Department of Computer Science, Electrical and Space Engineering, Luleå University of Technology, Sweden. Corresponding author's e-mail: gabcal@ltu.se}

\author[First]{Gabriele Calzolari}
\author[First]{Vidya Sumathy}
\author[First]{Christoforos Kanellakis}
\author[First]{George Nikolakopoulos} 

\address[First]{Department of Computer Science, Electrical and Space Engineering, Luleå University of Technology, Luleå, Sweden.}

\begin{abstract}                
This paper introduces a decentralized multi-agent reinforcement learning framework enabling structurally heterogeneous teams of agents to jointly discover and acquire randomly located targets in environments characterized by partial observability, communication constraints, and dynamic interactions. Each agent’s policy is trained with the Multi-Agent Proximal Policy Optimization algorithm and employs a Graph Attention Network encoder that integrates simulated range-sensing data with communication embeddings exchanged among neighboring agents, enabling context-aware decision-making from both local sensing and relational information. In particular, this work introduces a unified framework that integrates graph-based communication and trajectory-aware safety through safety filters. The architecture is supported by a structured reward formulation designed to encourage effective target discovery and acquisition, collision avoidance, and de-correlation between the agents' communication vectors by promoting informational orthogonality. The effectiveness of the proposed reward function is demonstrated through a comprehensive ablation study. Moreover, simulation results demonstrate safe, and stable task execution confirming the framework’s effectiveness.
\end{abstract}


\end{frontmatter}

\section{Introduction}
\label{sec:intro}

Heterogeneous multi-robot systems composed of Unmanned Aerial Vehicles (UAVs) and Unmanned Ground Vehicles (UGVs) offer complementary capabilities for complex missions such as surveillance, exploration, and search and rescue as highlighted by \cite{Shahar2025}. UAVs provide wide-area perception and rapid mobility, while UGVs contribute endurance, payload capacity, and ground-level sensing. Coordinating such diverse platforms in complex and communication-constrained environments requires the autonomous agents to make decentralized decisions under partial observability and distinct motion dynamics. These challenges have motivated the growing use of distributed Multi-Agent Reinforcement Learning (MARL) to enable multi-robot cooperation in uncertain environments as analyzed by \cite{Wang2022}. Recent advances have demonstrated the feasibility of learning decentralized control policies that scale to large teams and complex tasks as per \cite{NEURIPS2022_9c1535a0, kuba2022trustregionpolicyoptimisation, JING2025128729}. On-policy actor–critic methods such as the Multi-Agent Proximal Policy Optimization (MAPPO) algorithm have become strong baselines for cooperative tasks, while trust-region extensions, such as HAPPO and HATRPO derived from Heterogeneous-Agent Trust Region Learning (HATRL) proposed by \cite{JMLR:v25:23-0488}, improve training stability for heterogeneous agents. However, most of these frameworks ignore strong safety guarantees and inter-agent communication limitations, which restricts their deployment in physical robotic systems.

In parallel, graph-based neural architectures have emerged as an innate representation for relational reasoning and communication in multi-agent systems as presented by \cite{niu2021multi, hu2024learningmultiagentcommunicationgraph, ijcai2023p24}. By modeling agents as nodes and their interactions as edges, such methods learn both who to communicate with and what to share. Despite their success, these approaches typically neglect the safety-critical aspects of real-world robotics. Recent advances in safe reinforcement learning have expanded both the theoretical and algorithmic foundations of constrained policy optimization and safe exploration \cite{Gu2024review,wachi2024surveyconstraintformulationssafe}. These works formalize diverse safety specifications, ranging from cost-based and probabilistic constraints to state-wise safety requirements, and analyze their implications for policy learning and constraint satisfaction. Even-then, integration with heterogeneous MARL and communication learning remains limited.  

\subsection{Related works}

Multi-Agent Reinforcement Learning (MARL) has advanced cooperative control under uncertainty. In particular, the MAPPO framework investigated by \cite{NEURIPS2022_9c1535a0} shows that a simple actor–critic structure can yield robust policies with Centralized Training and Decentralized Execution (CTDE). Scalable MARL approaches aim to maintain learning efficiency and coordination as the number of agents grows. The work in \cite{CUI2025639} presents an MADDPG-based UAV–USV collaborative decision-making method that enhances visual relative-position estimation via CNNs, introduces tailored multi-constraint reward functions for heterogeneous cooperation, and achieves faster convergence and superior performance compared to traditional strategies. In this context, \cite{NEURIPS2024_fa76985f} introduces a scalable constrained policy optimization framework called Scalable MAPPO-Lagrangian (Scal-MAPPO-L), that enables safe and decentralized learning in large multi-agent systems by employing local $\kappa$-hop policy updates, and thereby mitigating the exponential growth of the joint state–action space while preserving cooperative performance. The work in \cite{OKAWA20231558} introduces Information-sharing Constrained Policy Optimization (IsCPO), a MARL framework that sequentially updates agents’ policies with shared surrogate-cost and divergence information to guarantee constraint satisfaction during learning and achieve suboptimal yet safe multi-agent policies in large-scale control tasks. \cite{guo2024mappo} further improves cooperative decision-making by using MAPPO-PIS, an intent-sharing–driven cooperative decision-making framework for Connected and Autonomous Vehicles (CAVs) that integrates Intention Generator Module (IGM) and the Safety Enhanced Module (SEM) correction to improve multi-agent policy learning, achieving superior safety, efficiency, and traffic performance in human–machine mixed merging scenarios. Although these frameworks are effective, many do not explicitly guarantee physical safety or enforce dynamic feasibility constraints. Our approach extends this foundation by incorporating safety filters for real-time, model-consistent safety within a MAPPO structure. Communication learning has emerged as a critical component of decentralized coordination. The Multi-Agent Graph-attentIon Communication (MAGIC) framework proposed by \cite{niu2021multi} employs a graph-attention communication protocol by jointly learning a Scheduler that determines when agents communicate and whom they address through a differentiable, attention-based dynamic graph encoder, and a Message Processor that applies Graph Attention Networks (GATs) over these dynamic graphs, enabling fully end-to-end training of communication structure and content. The Deep Hierarchical Communication Graph (DHCG) analyzed by \cite{ijcai2023p24} learns directed acyclic communication structures by optimizing graph topology end-to-end, enabling convergent and efficient information flow among agents. \cite{hu2024learningmultiagentcommunicationgraph} proposes CommFormer that reframes communication as a graph learning problem using transformer-based attention to infer sparse and adaptive communication topologies. Similarly, Agent Transformer Memory (ATM) \cite{NEURIPS2022_e1cf57f1} addresses partial observability in multi-agent RL by using a transformer-based working-memory mechanism to integrate factored entity observations with temporal information and to encode entity-specific action semantics. \cite{sun2025revisitingcommunicationefficiencymultiagent} formalized communication redundancy through a dimensional analysis of message embeddings and demonstrated that enforcing decorrelation within inter-agent representations significantly improves communication efficiency and cooperative performance in large-scale MARL systems. Comprehensive surveys \cite{Gronauer2022, huh2024multiagentreinforcementlearningcomprehensive} emphasize the importance of adaptive, efficient communication. Our framework introduces a communication dissimilarity regularization term in the reward function that explicitly penalizes correlated message vectors. 

The work in \cite{fan2024learn} introduce an offline-compatible safe experience reshaping mechanism that learns underlying constraint geometry, projects unsafe actions onto its tangent safe space, and trains policies exclusively on these safety-corrected experiences. A pixel-observation safe RL algorithm that jointly learns a low-dimensional latent dynamics model and a latent barrier-like safety function to encode state-wise constraints with unknown hazard regions, enabling simultaneous safe policy optimization that significantly reduces safety violations and accelerates safety convergence while maintaining competitive reward performance is proposed in \cite{zhan2024state}. To address the lack of a unified, practical methodology in SafeRL, where agents must optimize performance while minimizing risks of unintended harm, \cite{ji2024omnisafe} introduce a foundational, safety-centric framework that consolidates diverse RL algorithms into a cohesive, efficient research platform designed to streamline and accelerate progress in AI safety. Safe MARL frameworks \cite{GARG2024100945} and constrained optimization approaches \cite{NEURIPS2024_fa76985f} have begun exploring similar directions, but most remain centralized or homogeneous. Our work bridges this gap by embedding trajectory-aware safety filter for both holonomic (UAV) and differential-drive (UGV) agents directly into the decentralized MARL loop.

\begin{figure*}[t]
    \centering
    \includegraphics[width=0.9\linewidth]{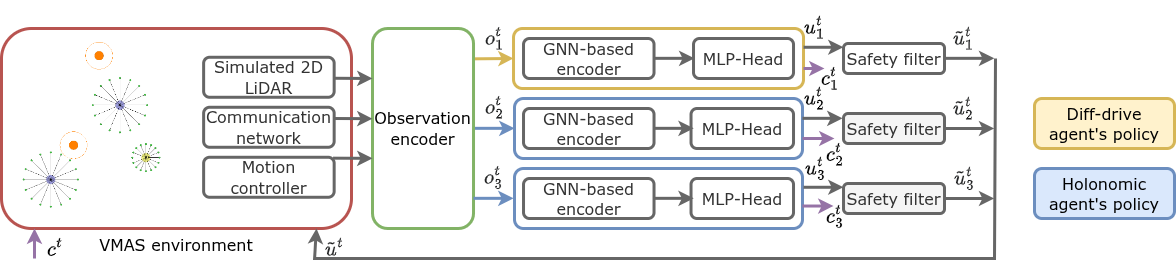}
    \caption{
        Illustration of the VMAS environment in which three agents are trained to locate and acquire randomly placed targets (orange). The diagram highlights the key components of each agent’s policy, including the GNN encoder, MLP heads, and safety filters.}
    \label{fig:vmas_concept}
\end{figure*}

\subsection{Our contribution}

To address these challenges, this paper proposes a decentralized MARL framework for structurally heterogeneous agents, such as UAV–UGV teams, that combines graph-based communication, safety filters that evaluate the feasibility of the trajectory induced by each proposed action and rescale it in the event of a predicted collision, and stable on-policy learning. Moreover, a communication orthogonality regularization enforces representational dissimilarity among agents’ message embeddings, mitigating interference and redundancy in shared communication channels. In particular, the main contributions are:

\begin{itemize}
\item A decentralized MARL architecture enabling coordination between structurally heterogeneous agents for collaborative target acquisition along with a communication orthogonality regularizer that promotes diverse messaging and robust multi-agent coordination.
\item A trajectory-based safety filter that enforces collision avoidance by predicting the state evolution induced by each proposed action and rescaling unsafe actions when a future violation is detected, ensuring that the executed control inputs remain within the admissible safe set.
\item Comprehensive ablation studies that quantify the contributions of goal-reaching, safety enforcement, and communication diversity to overall task performance.
\end{itemize}

\section{Methodology}

This section introduces the reinforcement learning framework in \ref{sbs:rl_framework}, the structure of the policy and critic networks in \ref{sbs:networks}, the proposed safety filters in \ref{sbs:safety_filters} and the reward function formulation in \ref{sbs:reward}.

\subsection{Reinforcement learning framework}
\label{sbs:rl_framework}

The partially-observable environment consists of a continuous 2D squared workspace of side length $d$ shown in Fig. \ref{fig:vmas_concept} that includes $n_t$ randomly distributed targets representing landmarks that $n_h$ holonomic agents (UAVs) with radius $r_h$ and $n_d$ differential-drive agents (UGVs) with radius $r_d$ have to explore and reach to.
We denote by $\mathcal{H}=\{1,\ldots,n_h\}$ and $\mathcal{D}=\{n_h+1,\ldots,n_h+n_d\}$ 
the sets of holonomic and differential-drive agents, respectively. 
The state of each agent $i$ is

\begin{equation}
\label{eq:agent_state}
    \mathbf{x}_i^t =
    \begin{cases}
        \big[\mathbf{p}_i^t,\ \mathbf{v}_i^t\big]^{\!\top} \in \mathbb{R}^4,
        & i \in \mathcal{H}, \\
        \big[\mathbf{p}_i^t,\ \theta_i^t,\ v_i^t,\ \omega_i^t\big]^{\!\top} \in \mathbb{R}^5,
        & i \in \mathcal{D},
    \end{cases}
\end{equation}

where $\mathbf{p}_i^t=[x_i^t,y_i^t]^{\!\top}$ is the position, 
$\mathbf{v}_i^t=[v_{x,i}^t,v_{y,i}^t]^{\!\top}$ is the velocity (holonomic), 
$\theta_i^t$ is the heading, $v_i^t$ the linear speed, and $\omega_i^t$ the angular speed of the agent $i$.
Moreover, all agents are equipped with a simulated range sensor based on ray casting, which provides $n_l$ range measurements indicating detected targets or the maximum sensing range, denoted $r_h^l$ for holonomic agents and $r_d^l$ for differential-drive agents.
At the beginning of each episode, the agents have no prior information about the environment or target locations. A target $g \in \mathcal{G}$ is deemed considered covered when at least one agent lies within the distance threshold $\rho_{\text{cov}}$, as defined in Eq.~\eqref{eq:target_cov}.
\begin{equation}
\label{eq:target_cov}
    \min_{i \in \mathcal{H} \cup \mathcal{D}}
    \big\| \mathbf{p}^t_i - \mathbf{p}^t_g \big\|_2
    \;\le\;
    \rho_{\text{cov}}
\end{equation}
To explore the environment and find the targets, at each step $t$, each agent $i$ outputs an action $a_i^t$ composed by a continuous movement action $\bm{u}_i^t \in \mathbb{R}^2$ subject to the agent's dynamics and a communication vector $\bm{c}^t_i \in \mathbb{R}^{d_c}$. Agents are allowed to exchange information with agents within a finite communication radius $r_c$. These actions are computed by the policies based on some observations received from the environment at the beginning of the step. In particular, each agent $i$ receives an observation vector $\mathbf{o}_i^t$ that combines its sensor measurements, the communication vector, and selected attributes of its own state as per Eq. \eqref{eq:obs_vector}.
\begin{equation}
\label{eq:obs_vector}
    \mathbf{o}_i^t =
    \begin{cases}
        \big( l_i^t,\ \mathbf{c}_i^t,\ \mathbf{v}_i^t \big), & i \in \mathcal{H}, \\
        \big( l_i^t,\ \mathbf{c}_i^t,\ \mathbf{v}_i^t,\ \theta_i^t,\ \omega_i^t \big), & i \in \mathcal{D},
    \end{cases}
\end{equation}
where $l_i^t$ denotes the range-sensor readings, $\mathbf{c}_i^t$ the communication vector, 
$\mathbf{v}_i^t$ the agent's velocity, $\theta_i^t$ the heading, and $\omega_i^t$ the angular velocity. Upon selecting an action $\bm{u}_i^t$, a safety filter $\Gamma$ is applied, with its formulation defined in \ref{sbs:safety_filters}. The filter enforces safety constraints by projecting the nominal control input $\mathbf{u}_i^{t}$ generated by the policy onto the admissible action set, resulting in the safe action $\tilde{\mathbf{u}}_i^t$. Thus ensuring that the executed command does not lead to collisions with other agents.

Holonomic agents are fully actuated in the plane and use a two-dimensional safe control input $\tilde{\mathbf{u}}_i^t
 = [\,\tilde{f}_x, \tilde{f}_y\,]^\top$, where each force component satisfies $\tilde{f}_x, \tilde{f}_y \in [-u_{\max}, u_{\max}]$. The safe control input directly affects the discrete-time dynamics, which can be written in state-update form as per Eq. \eqref{eq:holo_dynamics}.
\begin{equation}
\label{eq:holo_dynamics}
    \mathbf{x}_i^{t+1}
    =
    \begin{bmatrix}
        \mathbf{p}_i^{t+1} \\
        \mathbf{v}_i^{t+1}
    \end{bmatrix}
    =
    \begin{bmatrix}
        \mathbf{p}_i^{t} + \mathbf{v}_i^{t}\,\Delta t \\
        \mathbf{v}_i^{t} + \Big( \tfrac{1}{m_i}\tilde{\mathbf{u}}_i^t - c_d\,\mathbf{v}_i^{t} \Big)\Delta t
    \end{bmatrix}
\end{equation}

where $m$ denotes the agent's mass, $\Delta t$ is the environment's time-step, and $c_d \!\ge\! 0$ represents the linear drag coefficient. Differential-drive agents evolve according to unicycle kinematics integrated via a fourth-order Runge-Kutta (RK4) scheme. Their safe control action is given by $\tilde{\mathbf{u}}_i^t = [\, \tilde{v}_i^{t},\ \tilde{\omega}_i^{t} \,]^\top$, where the linear and angular velocities satisfy $\tilde{v}_i^{t},\, \tilde{\omega}_i^{t} \in [-u_{\max},\, u_{\max}]$.

\subsection{Policy and critic networks}
\label{sbs:networks}

Each agent’s policy is structured as a Graph Neural Network (GNN) encoder followed by a Multi-Layer Perceptron (MLP) output head. 
The first stage of each agent’s policy is represented by a graph neural network (GNN) defined over a dynamic, position-dependent graph 
$\mathcal{G}_t = (\mathcal{V}, \mathcal{E}_t)$, 
where each node $v_i \in \mathcal{V}$ corresponds to an agent at time $t$. 
The edge set $\mathcal{E}_t$ is constructed from the agents' spatial configuration, such that 
an edge exists between two agents if their Euclidean distance is within the predefined 
communication range $r_c$ according to Eq. \eqref{eq:graph_edges}.
\begin{equation}
\label{eq:graph_edges}
    \mathcal{E}_t = \{ (v_i, v_j) \mid \| \mathbf{p}^t_i - \mathbf{p}^t_j \|_2 \le r_c, \, i \neq j \}
\end{equation}
Each node feature vector contains the local observation of the corresponding agent. The GNN then constructs edge features by computing pairwise relational quantities, specifically the relative position $\mathbf{p}^t_i - \mathbf{p}^t_j$, the inter-agent distance $\|\mathbf{p}^t_i - \mathbf{p}^t_j\|_2$, and the relative velocity $\mathbf{v}^t_i - \mathbf{v}^t_j$. Message passing within the GNN is implemented using the GATv2Conv operator, proposed by \cite{brody2022attentivegraphattentionnetworks}, which performs attention-weighted aggregation over neighboring nodes. 
After graph aggregation, each agent passes its embedding through a policy head conditioned on its dynamic model where each MLP consists of two hidden layers with 256 units and ELU activations followed by a linear output layer matching the action dimension. 
Furthermore, the critic is implemented using a DeepSets architecture proposed by \cite{zaheer2018deepsets}. The joint observation $ \mathbf{o}^t
    =
    \big(
        \{ \mathbf{o}_i^t \}_{i \in \mathcal{H}},
        \{ \mathbf{o}_i^t \}_{i \in \mathcal{D}}
    \big)$ is first embedded through a local encoder $\phi(\bm{o}^t)$, and the global value is obtained by symmetric aggregation according to Eq. \eqref{eq:deepsets}.
\begin{equation}
V(\bm{s}) = 
\rho\!\left(
\frac{1}{N_a}
\sum_{i=1}^{N_a}
\phi(\bm{o}^t)
\right),
\label{eq:deepsets}
\end{equation}
where $\phi(\cdot)$ and $\rho(\cdot)$ are mappings realized as MLPs with ELU activations. In our setup, $\phi$ consists of two hidden layers of 128 neurons and $\rho$ of two layers with 256 neurons.

\subsection{Safety filters}
\label{sbs:safety_filters}

To guarantee real-time safety, each agent’s control input is projected onto the closest admissible control that satisfies a prescribed set of safety constraints, ensuring that the predicted trajectory remains collision-free. In particular, $d_{\text{safe}}$ is the minimum required inter-agent separation to avoid collision.

\subsubsection{Holonomic agents}

To guarantee collision avoidance under uncertain neighbor motion, it is proposed a safety filter that selects the closest feasible control input
to the nominal force while ensuring that the agent's trajectory
remains collision-free over the entire prediction horizon.
For any candidate scaled control input $\tilde{\mathbf{u}}_i(\alpha) = \alpha\,\mathbf{u}_i^t$
where $\alpha \in \mathcal{A} = \{1.0,\,0.8,\,0.6,\,0.4,\,0.25,\,0.1,\,0.0\},
$ the safety filter evaluates the feasibility of the resulting trajectory by propagating the agent's motion under 
$\tilde{\mathbf{u}}_i(\alpha)$ and verifying that it remains outside the maximum possible occupancy regions of all nearby agents. 
Among all feasible candidates, the control input corresponding to the largest admissible scaling factor is selected. In particular, the ego agent's trajectory is approximated over 
$\tau \in [0,\Delta t]$ as
\begin{equation}
\begin{aligned}
    \mathbf{p}_i^{\,t+\tau}(\alpha)
    &= 
    \mathbf{p}_i^{\,t}
    +
    \Bigl[
        \mathbf{v}_i^{\,t}
        + \tfrac{\Delta t}{2m_i}\,\mathbf{u}_i(\alpha)
    \Bigr]\tau
    \\
    \text{s.t.}\quad
    &
    \bigl\lVert 
        \mathbf{p}_i^{\,t+\tau}(\alpha)
        - \mathbf{p}_j^{\,t+\tau}
    \bigr\rVert_2
    \;\ge\;
    d_{\mathrm{safe}}.
\end{aligned}
\label{eq:ego_traj}
\end{equation}

Each neighboring holonomic agent $j$ possesses bounded per-axis
velocities 
$\mathbf{v}^t_j \in [-v_{\max}, v_{\max}]$.
This induces a worst-case reachable set at time $t+\Delta t$ described by the
axis-aligned rectangle
\begin{equation}
    \mathcal{R}^t_j 
    = \big\{\, \mathbf{p} \in \mathbb{R}^2 
        \;\big|\;
        \lvert \mathbf{p}-\mathbf{C}^t_j \rvert_{\infty} 
        \le \mathbf{H}_j
      \big\},
\end{equation}
where the center and half-size are given by
\begin{equation}
    \mathbf{C}^t_j = 
        \mathbf{p}^t_j + \tfrac{1}{2}
        (\mathbf{v}_{\min}+\mathbf{v}_{\max}) \Delta t,
    \qquad
    \mathbf{H}_j = 
        \tfrac{1}{2}
        (\mathbf{v}_{\max}-\mathbf{v}_{\min}) \Delta t.
\end{equation}
To account for the circular bodies of the ego and neighbor agents, this set is
inflated by the sum of agents' radii $r_i + r_j$ in the $L_\infty$ metric,
\begin{equation}
    \mathcal{R}_j^+
    =
    \big\{\, \mathbf{x} \in \mathbb{R}^2
    \;\big|\;
    \lvert \mathbf{x}-\mathbf{C}^t_j\rvert_{\infty}
    \le 
    \mathbf{H}_j + (r_i+r_j++d_{\mathrm{safe}})\mathbf{1}
    \big\}.
\end{equation}

In addition to holonomic neighbors, the environment contains differential-drive agents with maximum linear speed $v_{\max}$.
Over a short horizon $\Delta t$, the exact unicycle reachable set is a
nonlinear subset of a circular sector whose angular width is bounded by
$\omega^{\max}\Delta t$.
To enable efficient continuous-time collision checking, a conservative over-approximation is adopted by assigning to each diff-drive agent an isotropic velocity bound $\big\|\dot{\mathbf{p}}^t_j\big\|_2 \le v^{\max}$ which implies that its position at time $t+\Delta t$ belongs to the disc
\begin{equation}
    \mathcal{D}^t_j =
    \big\{\, \mathbf{x} \in \mathbb{R}^2 
    \;\big| \;
    \big\|\mathbf{x} - \mathbf{p}^t_j \big\|_2
    \le v^{\max}\Delta t
    \big\}.
\end{equation}
This disc is inflated by the sum of agents' radii and an additional safety
distance $d_{\mathrm{safe}}$, yielding the conservative collision set
\begin{equation}
    \mathcal{D}_j^+
    =
    \big\{\, \mathbf{x} \in \mathbb{R}^2 \;\big|\;
    \big\|\mathbf{x} - \mathbf{p}^t_j \big\|_2
    \le v^{\max}\Delta t + (r_i+r_j+d_{\mathrm{safe}})
    \big\}.
\end{equation}

Finally, it is selected the $\alpha$ that satisfies Eq. \eqref{eq:alpha_opt}.

\begin{equation}
\begin{aligned}
    \max_{\alpha\in \mathcal{A}} \quad & \alpha \\[2mm]
    \text{s.t.} \quad 
    & \mathbf{p}_i(t+\tau;\alpha)\;\notin\;\mathcal{C},
      \qquad \forall\,\tau\in[0,\Delta t].
\end{aligned}
\label{eq:alpha_opt}
\end{equation}
where 
\begin{equation}
    \mathcal{C}
    \;=\;
    \Bigl(\,\bigcup_{j\in H}\mathcal{R}_j^{+}\Bigr)
    \;\cup\;
    \Bigl(\,\bigcup_{j\in D}\mathcal{D}_j^{+}\Bigr).
    \label{eq:collision_set}
\end{equation}

\subsubsection{Differential drive agent}

Similar considerations are done for the definition of the safety filter used by the differential drive agent but in this case the forward evolution under $u(\alpha)$ is approximated using a fourth-order
Runge--Kutta (RK4) integrator,
and the interval $[0,\Delta t]$ is subdivided into 5 uniform substeps to produce intermediate states $\{x_k(\alpha)\}_{k=1}^5$. Each intermediate position $p_k(\alpha)\in\mathbb{R}^2$ is subjected to collision checks in a way analogous to the one discussed before for the holonomic agent.

\begin{figure*}[h!]
    \centering
    \begin{subfigure}[b]{0.33\textwidth}
        \centering
        \includegraphics[width=\textwidth]{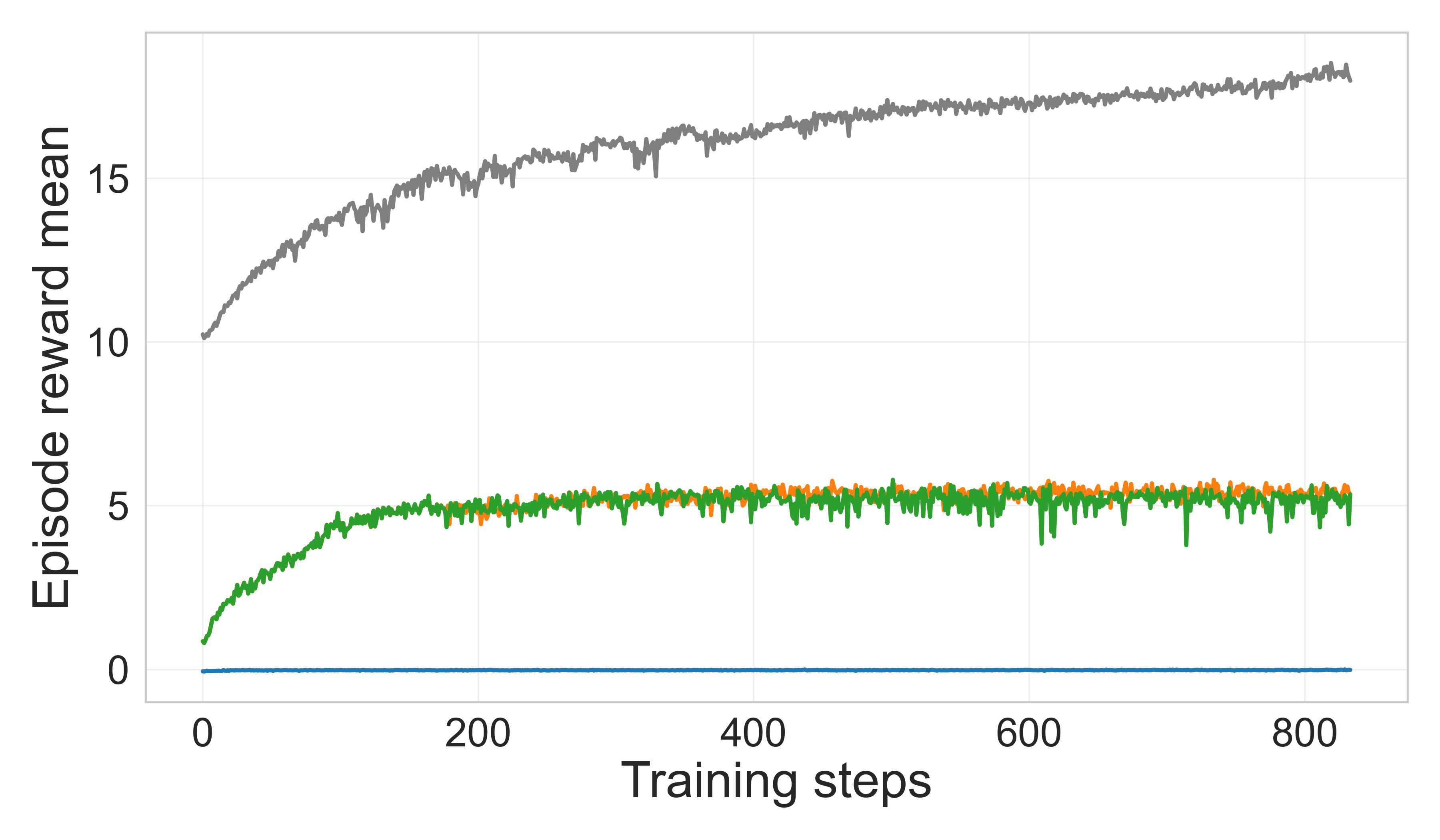}
        \caption{}
        \label{fig:total_episode_reward}
    \end{subfigure}\hfill
    \begin{subfigure}[b]{0.33\textwidth}
        \centering
        \includegraphics[width=\textwidth]{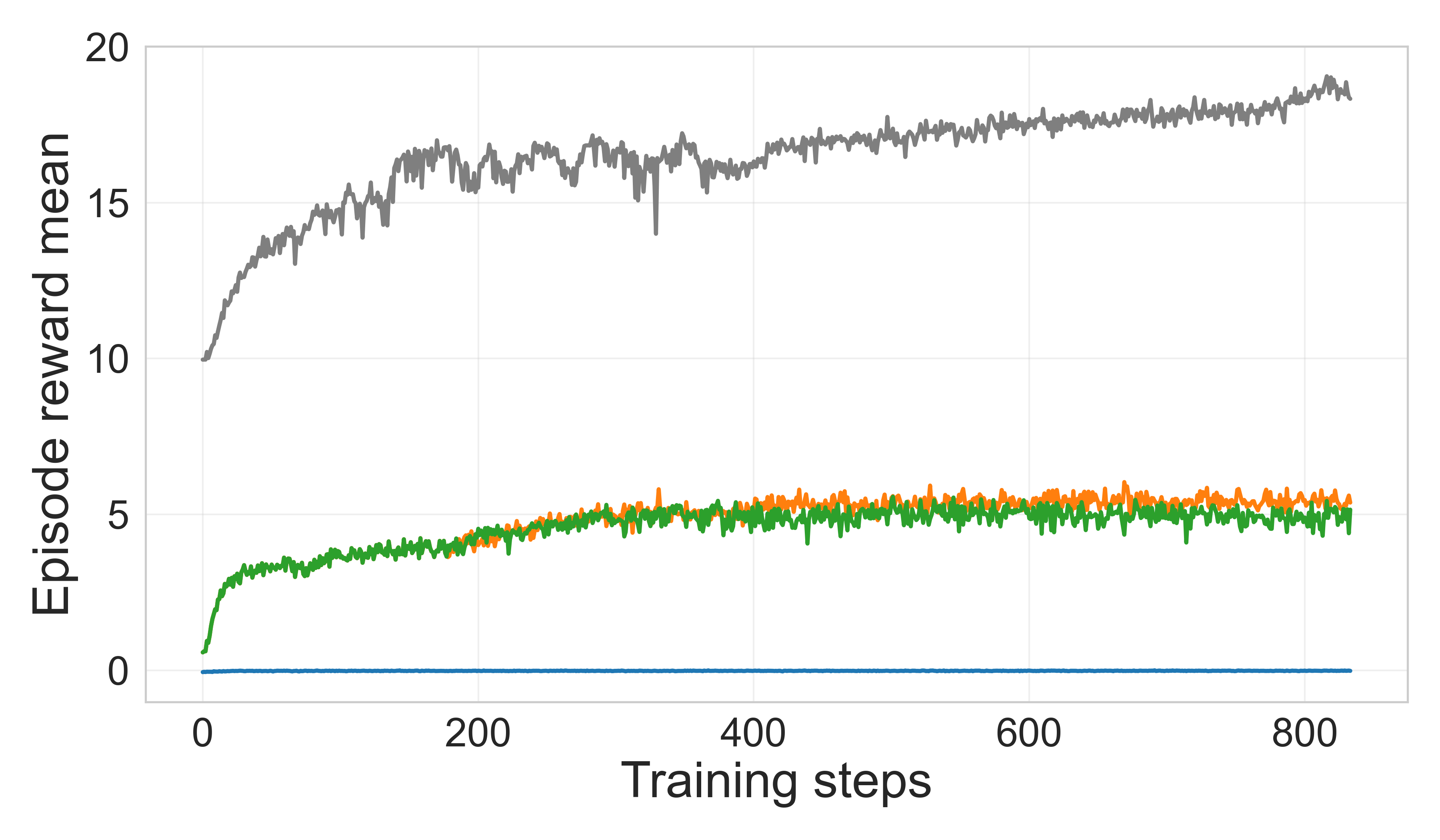}
        \caption{}
        \label{fig:holo_episode_reward}
    \end{subfigure}\hfill
    \begin{subfigure}[b]{0.33\textwidth}
        \centering
        \includegraphics[width=\textwidth]{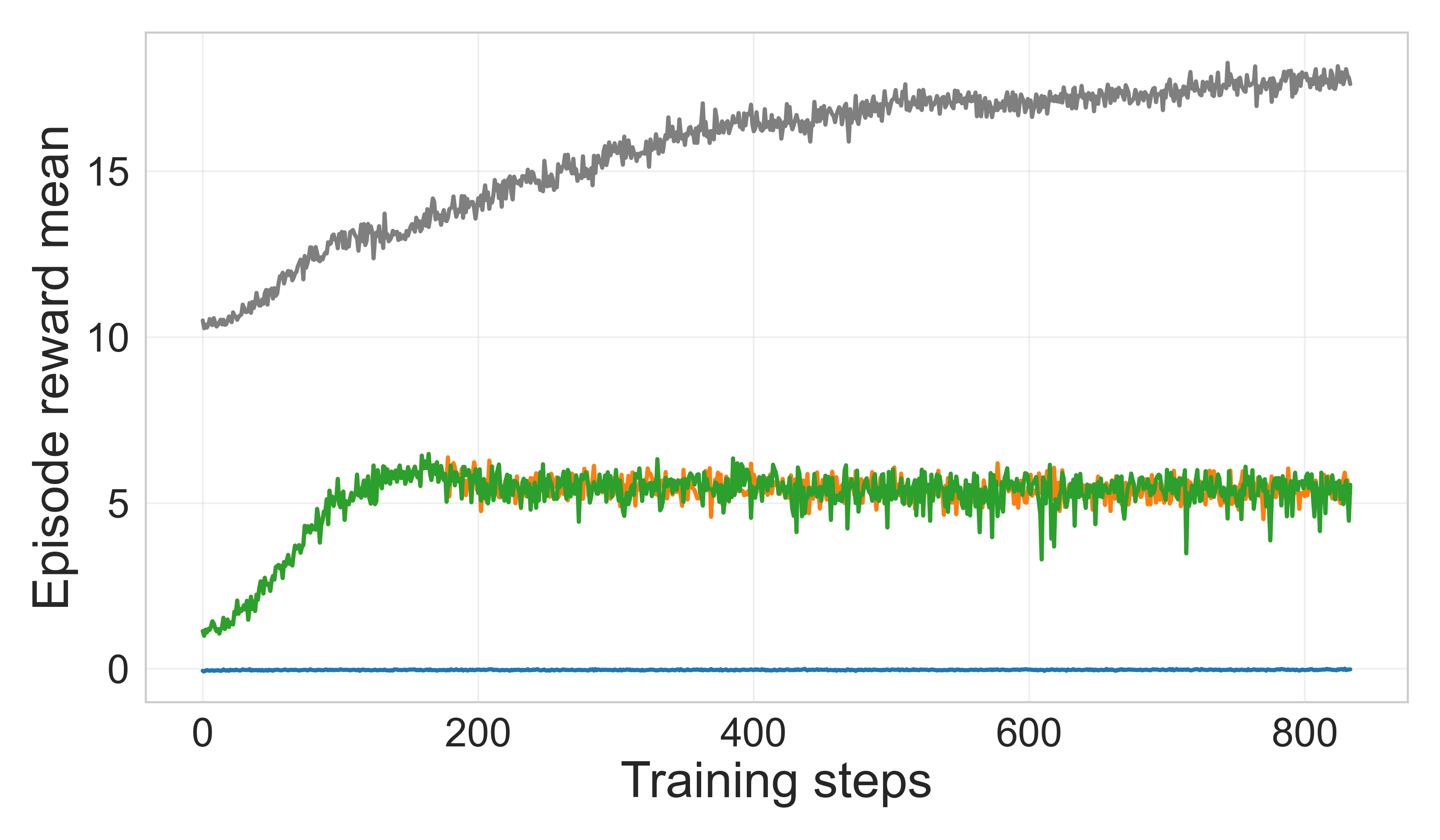}
        \caption{}
        \label{fig:diff_drive_episode_reward}
    \end{subfigure}

    \caption{Comparison of the mean episode reward trajectories for the full multi-agent system (a), holonomic agents (b), and diff-drive agents (c) under different reward shaping configurations. Reward schemes are color-coded as follows: R1 (blue), R2 (orange), R3 (green), and R4 (grey).}
    \label{fig:episode_reward_trajectories}
\end{figure*}

\subsection{Reward function formulation}
\label{sbs:reward}

At each step $t$, agent $i$ receives a scalar reward $r_i^t$ expressed by Eq. \eqref{eq:proposed_reward}.
\begin{equation}
\label{eq:proposed_reward}
r_i^t \;=\; \bm{w}^\top \bm{r}_i^t
\;=\;
\begin{bmatrix}
w_{\text{dist}} & w_{\text{goal}} & w_{\text{coll}} & w_{\text{comm}}
\end{bmatrix}
\begin{bmatrix}
r_i^{\text{dist}}(t) \\[3pt]
r_i^{\text{goal}}(t) \\[3pt]
r_i^{\text{coll}}(t) \\[3pt]
r_i^{\text{comm}}(t)
\end{bmatrix},
\end{equation}
where $\bm{w}\!\in\!\mathbb{R}^4$ contains the reward weights and $\bm{r}_i^t\!\in\!\mathbb{R}^4$ the corresponding reward terms. These last terms capture progress toward targets, goal completion, collision avoidance, and communication diversity. The distance-based reward $r_i^{\text{dist}}(t)$ encourages agents to approach their nearest target and is defined as
\begin{equation}
\label{eq:r-dist}
r_i^{\text{dist}}(t) =
\min_{g \in \mathcal{G}} \| \mathbf{p}_i^t - \mathbf{p}_g^t \|
-
\min_{g \in \mathcal{G}} \| \mathbf{p}_i^{\,t-1} - \mathbf{p}^{t-1}_g \|.
\end{equation}
Moreover, when the agent covers a target, a fixed positive reward $r_{\text{goal}}$ is assigned to $r_i^{\text{goal}}(t)$. To discourage unsafe proximity to other agents, we define $r_i^{\text{coll}}(t) = r_{\text{coll}}$ whenever a collision happens. Furthermore, each agent produces a message vector $c_i^t \in \mathbb{R}^{d_c}$, and message diversity is encouraged via Eq. \eqref{eq:r-comm}.
\begin{equation}
\label{eq:r-comm}
r_i^{\text{comm}}(t)
\;=\;
\sum_{j\neq i}\!\left(1 - \Gamma^2\!\big(c_i^t, c_j^t\big)\right),
\quad
\Gamma(c_i^t,c_j^t)=\frac{\langle c_i^t, c_j^t \rangle}
{\|c_i^t\|\,\|c^t_j\|},
\end{equation}

To assess the contribution of the individual reward components, it is performed an ablation study using the progressively structured reward functions given by the reward terms vectors defined in Eq. \ref{eq:abl}. The analysis encompasses both the performance metrics recorded during policy training and the outcomes obtained in simulation over randomly generated environments, as reported in \ref{sbs:training_results} and \ref{sbs:simulation_results}.

\begin{equation}
\label{eq:abl}
\begin{aligned}
\textbf{R1:}\quad 
\bm{r}_i^{(1)}(t) &= [\,r_i^{\text{dist}}(t),\ 0,\ 0,\ 0\,]^\top,
\\[4pt]
\textbf{R2:}\quad 
\bm{r}_i^{(2)}(t) &= [\,r_i^{\text{dist}}(t),\ r_i^{\text{goal}}(t),\ 0,\ 0\,]^\top,
\\[4pt]
\textbf{R3:}\quad 
\bm{r}_i^{(3)}(t) &= [\,r_i^{\text{dist}}(t),\ r_i^{\text{goal}}(t),\ r_i^{\text{coll}}(t),\ 0\,]^\top,
\\[4pt]
\textbf{R4:}\quad 
\bm{r}_i^{(4)}(t) &= [\,r_i^{\text{dist}}(t),\ r_i^{\text{goal}}(t),\ r_i^{\text{coll}}(t),\ r_i^{\text{comm}}(t)\,]^\top.
\end{aligned}
\end{equation}

\section{Results}

This section presents the training setup in \ref{sbs:rl_framework} and the training and simulation results in \ref{sbs:training_results} and \ref{sbs:simulation_results}, respectively.

\subsection{Training}
\label{sbs:training}

The agents are trained within a custom environment built on the VMAS simulator proposed by \cite{bettini2022vmasvectorizedmultiagentsimulator}, configured according to the parameters reported in Table \ref{tab:env_rew_params}. Moreover, the training is performed using the BenchMARL framework proposed by \cite{bettini2024benchmarl}, leveraging its implementation of the Multi-Agent Proximal Policy Optimization (MAPPO) algorithm. In particular, MAPPO is an on-policy actor-critic method adhering to the centralized training and decentralized execution (CTDE) paradigm, whereby each agent learns an individual policy, while a shared centralized critic is optimized using joint observations to stabilize multi-agent learning.

\begin{table}[t]
\centering
\caption{Key environment and reward function parameters}
\label{tab:env_rew_params}
\begin{tabular}{cccc}
\toprule
Symbol & Value & Symbol & Value \\
\midrule
$n_d, n_h, n_t, d$ & $1, 2, 3, 10m$ & $w_{\text{dist}}$ & $1.0$ \\
$r_d, r_h, v_{\max}$ & $0.5m, 0.5m, 10m/s$ & $w_{\text{goal}}$ & $1.0$ \\
$n_l, r^l_d, r^l_h$ & $16, 1.5m, 3.0m$ & $w_{\text{coll}}$ & $1.0$ \\
$d_{\text{safe}}, \rho_{\text{cov}}$ & $0.05m, 1.5m$ & $w_{\text{comm}}$ & $0.1$ \\
$m, u_{\max}, \Delta t, c_d$ & $1kg,1,0.1s, 0.25$ & $r_{\text{goal}}$ & $10.0$ \\
$d_c, r_c$ & $16, 4.5m$ & $r_{\text{coll}}$ & $-8.0$ \\
\bottomrule
\end{tabular}
\end{table} 

\begin{figure*}[h!]
    \centering
    \begin{subfigure}[b]{0.33\textwidth}
        \centering
        \includegraphics[width=\textwidth]{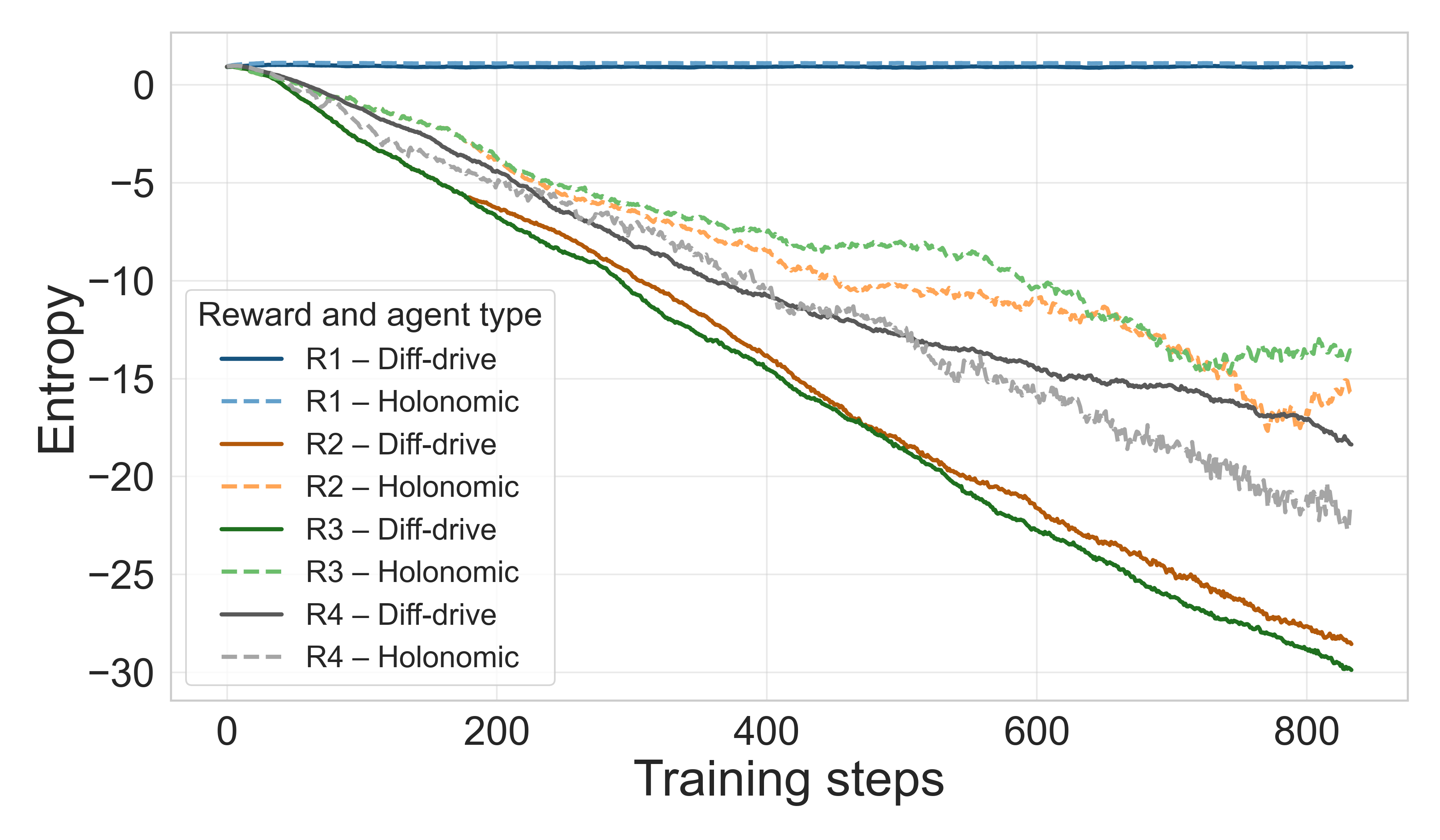}
        \caption{}
        \label{fig:entropy_agents}
    \end{subfigure}\hfill
    \begin{subfigure}[b]{0.33\textwidth}
        \centering
        \includegraphics[width=\textwidth]{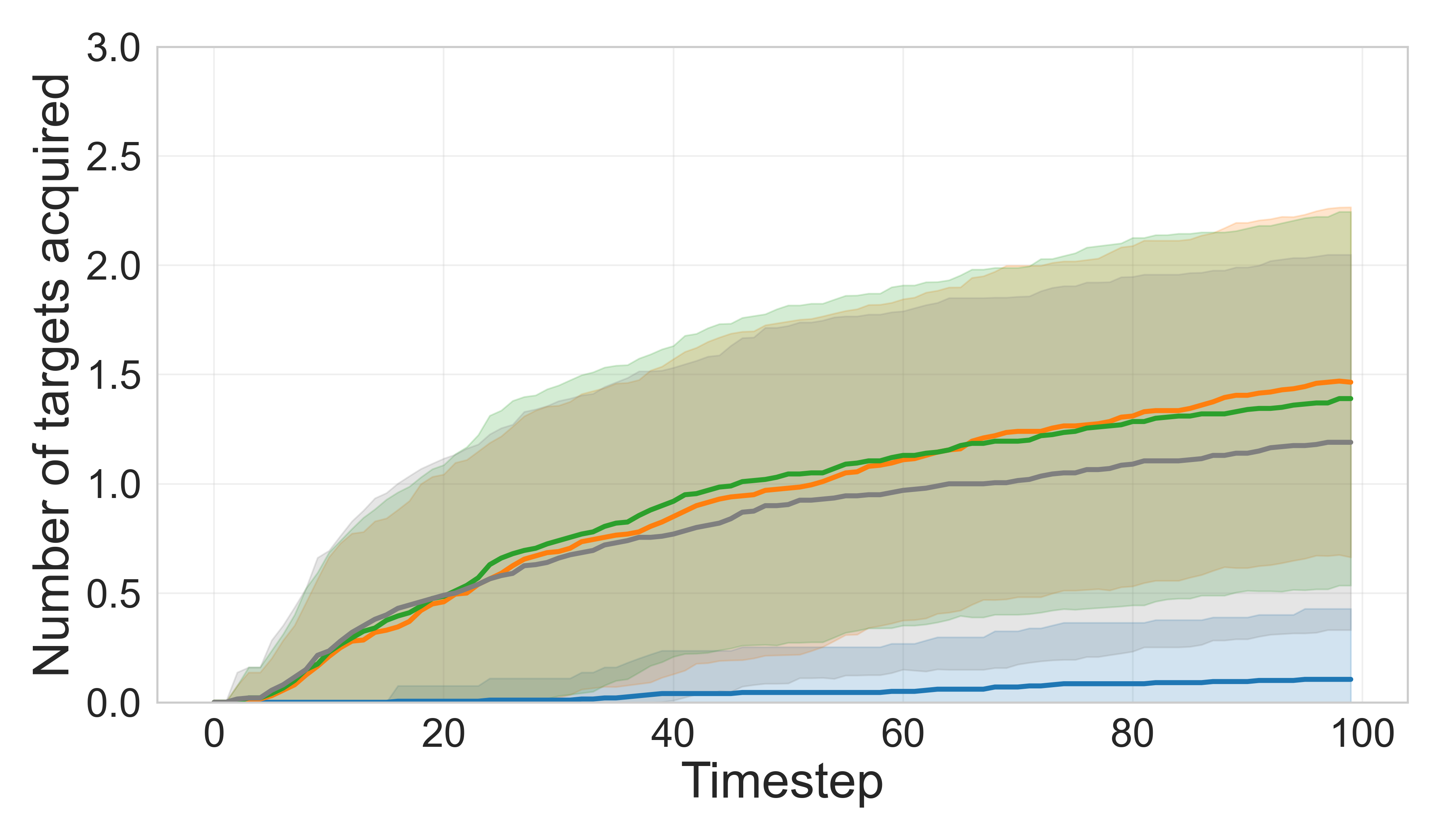}
        \caption{}
        \label{fig:traj_target_acq}
    \end{subfigure}\hfill
    \begin{subfigure}[b]{0.33\textwidth}
        \centering
        \includegraphics[width=\textwidth]{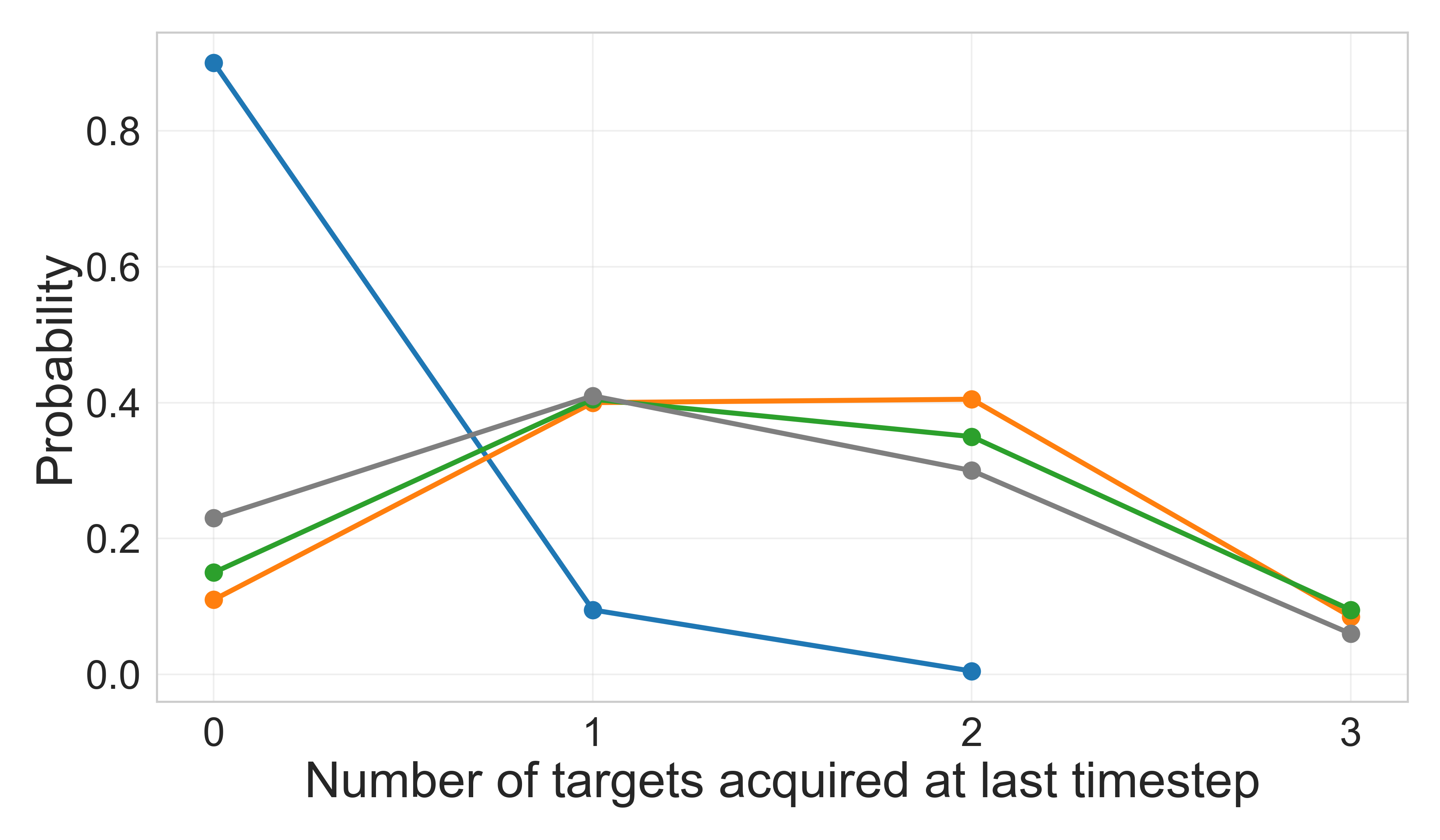}
        \caption{}
        \label{fig:target_acq_last}
    \end{subfigure}\hfill
    \caption{(a) Entropy evolution for holonomic and diff-drive agents under the four reward
configurations. (b) Mean number of acquired targets over time with standard deviation
bands. (c) Probability mass function of the number of targets acquired at the final
timestep. Colors denote reward schemes: R1 (blue), R2 (orange), R3 (green), and R4 (grey).}

    \label{fig:role_target}
\end{figure*}

\begin{figure*}[]
    \centering
    \includegraphics[width=\linewidth]{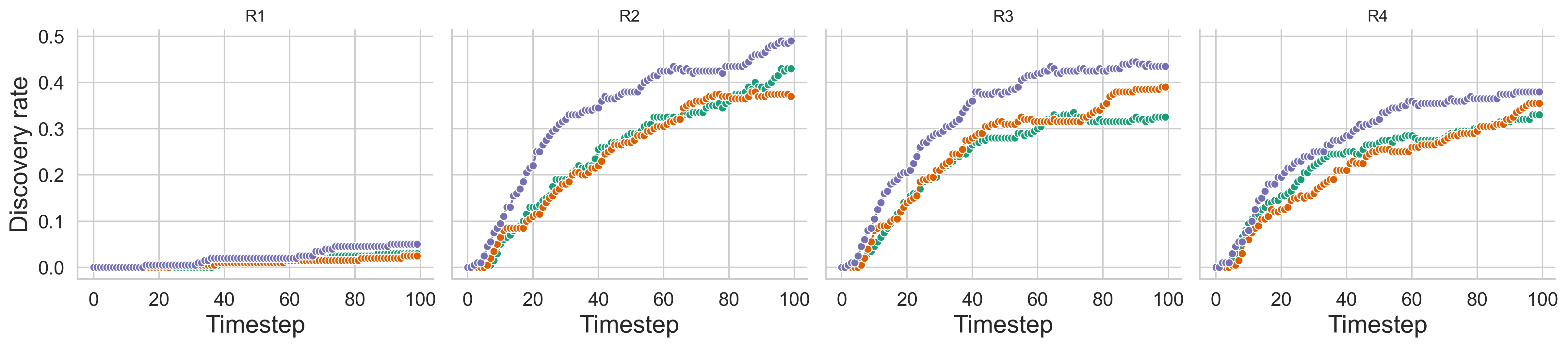}
    \caption{Per-agent target discovery over time for the four reward configurations (R1–R4). Each subplot shows the mean discovery rate across simulations, with colored curves indicating the diff-drive agent (green), holonomic agent 1 (orange), and holonomic agent 2 (purple).}
    \label{fig:agent_distr}
\end{figure*}

\subsection{Training results}
\label{sbs:training_results}
The agent policies have been trained for 834 optimization steps in 600 parallel environments using MAPPO. Figure \ref{fig:total_episode_reward} shows the
evolution of the mean episode reward collected during training across the four reward configurations. Policies trained under R2 and R3 exhibit clear and
stable convergence, following highly similar learning trajectories. In contrast, R1 remains close to zero throughout training, indicating that this reward
configuration fails to provide a sufficiently informative learning signal. R4 achieves the highest episode rewards and shows evidence of convergence,
although with noticeably higher variance. A similar behaviour is observed when analysing the mean episode reward per agent type, as shown in Figures~\ref{fig:holo_episode_reward} and
\ref{fig:diff_drive_episode_reward}. For both R2 and R3, the differential-drive agent consistently achieves slightly higher rewards than the holonomic agents,
whereas under R4 the holonomic agents marginally outperform the differential-drive one. The close correspondence between the per-agent reward curves across all reward configurations suggests that the learned policies support effective collaborative task execution, without inducing a pronounced asymmetric
workload among the heterogeneous agents. Figure \ref{fig:entropy_agents} reports the entropy evolution of the learned policies, which provides insight into the exploration--exploitation dynamics during training. Under R2, R3, and R4, entropy decays rapidly, indicating that the agents progressively acquire stable and consistent behaviours. Notably, for
R2 and R3 the holonomic agents retain higher entropy for longer than the differential-drive agent, suggesting a degree of behavioural specialization.
Conversely, R4 yields nearly identical entropy profiles across agent types, indicating more symmetric role emergence. In contrast, R1 maintains high
entropy throughout training, reflecting persistent policy randomness and the absence of meaningful behavioural convergence.

\subsection{Simulation results}
\label{sbs:simulation_results}
The trained policies have been further evaluated on 200 simulated environments,
each executed for 100 timesteps, using scenarios consistent with those employed
during training. Figure \ref{fig:traj_target_acq} reports the mean number of
targets acquired over time and provides an overall indication of task
execution efficiency. Policies trained with R1 exhibit minimal progress,
confirming their inability to explore the environment effectively and to
locate targets. In contrast, the policies trained with R2, R3, and R4
successfully acquire targets within the available time horizon. The steep
initial increase observed for the latter configurations indicates that the
agents have learned meaningful acquisition behaviours. However, the
considerable variance associated with R2 and R3 suggests reduced robustness
and a stronger dependence on the specific simulation instance. These observations are corroborated in
Figure \ref{fig:target_acq_last}, which shows the probability mass function of
the number of targets acquired at the final timestep. Policies obtained with R2,
R3, and R4 typically manage to acquire one or two targets, whereas R1 fails to
achieve comparable performance in most evaluation runs. To assess the distribution of workload across the heterogeneous agents, Figure \ref{fig:agent_distr} presents the per-agent target discovery dynamics.
R4 yields policies in which the discovery trajectories of the diff-drive and
holonomic agents are closely aligned, indicating a more balanced workload and
thus more symmetric collaboration. Conversely, R2 and R3 lead to more
pronounced discrepancies between agents, revealing a degree of specialization
and a less uniform allocation of the acquisition effort. Among these,
R3 exhibits more stable behaviour than R2, as reflected by the lower variance
in entropy during training and by the clearer role differentiation observed for
the two holonomic agents relative to the diff-drive agent. Across all evaluation metrics, policies trained under R1 demonstrate markedly
inferior performance. They fail to develop reliable exploration or acquisition
strategies and consistently underperform relative to the other reward
configurations.

\section{Conclusion}
\label{sec:conclusion}

This work presents a decentralized multi-agent reinforcement learning framework for heterogeneous robotic teams for target acquisition in unknown environments, combining graph-based policies with safety filters to ensure collision-free coordination. The simulation results demonstrated stable learning, effective cooperation, and balanced workload distribution across agents under suitable reward shaping. Moreover, ablation studies confirmed the importance of goal-oriented, collision-avoidance, and communication-related reward terms. Future work will improve the performances in the target acquisition, investigate the efficacy of the framework with larger teams, and real-world deployment of the learned policies.

\begin{ack}
The computations were enabled by resources provided by the National Academic Infrastructure for Supercomputing in Sweden (NAISS) and Chalmers e-Science Centre at C3SE, partially funded by the Swedish Research Council through grant agreement no. 2022-06725. Computations were performed on the Alvis cluster.
\end{ack}

\bibliography{ifacconf}             
                                                   







\end{document}